# LB-CNN: An Open Source Framework for Fast Training of Light Binary Convolutional Neural Networks using Chainer and Cupy

Radu Dogaru
*Dept. of Applied and Information Engineering*
*University "Politehnica" of Bucharest*
Bucharest, Romania
radu.dogaru@upb.ro

Ioana Dogaru
*Dept. of Applied and Information Engineering*
*University "Politehnica" of Bucharest*
Bucharest, Romania
ioana.dogaru@upb.ro

*Abstract*— Light binary convolutional neural networks (LB-CNN) are particularly useful when implemented in low-energy computing platforms as required in many industrial applications. Herein, a framework for optimizing compact LB-CNN is introduced and its effectiveness is evaluated. The framework is freely available and may run on free-access cloud platforms, thus requiring no major investments. The optimized model is saved in the standardized .h5 format and can be used as input to specialized tools for further deployment into specific technologies, thus enabling the rapid development of various intelligent image sensors. The main ingredient in accelerating the optimization of our model, particularly the selection of binary convolution kernels, is the Chainer/Cupy machine learning library offering significant speed-ups for training the output layer as an extreme-learning machine. Additional training of the output layer using Keras/Tensorflow is included, as it allows an increase in accuracy. Results for widely used datasets including MNIST, GTSRB, ORL, VGG show very good compromise between accuracy and complexity. Particularly, for face recognition problems a carefully optimized LB-CNN model provides up to 100% accuracies. Such TinyML solutions are well suited for industrial applications requiring image recognition with low energy consumption.

## I. INTRODUCTION

Image recognition is a widely recognized problem. In recent years edge computing solutions emerge where nodes are low-complexity sensors [1][2]. As recognized in [3] "industry quest for ultra-low-power neural networks is just at the beginning", and recently several industrial applications [4][5] were reported as successful cases for using convolutional neural networks (CNNs). With the wide development of machine learning solutions for computer vision problems, two main problems arise: i) the improvement of the functional performance, often given as the "accuracy" i.e. the fraction of correctly recognized samples in a test set; ii) low-complexity and low-power implementation solutions, while keeping good accuracy (while accepting some sacrifice in performance when compared with the most accurate model). Our work fits in the second direction, providing a framework for fast training and optimizing the model parameters of simple convolutional neural network (CNN) model, denoted here LB-CNN. In this model, binary quantization of kernels in the convolution layers is performed, enabling fast and compact implementation. A recent overview of binary CNNs [6] reveals a wide range of solutions, many providing also computational frameworks with open-source code. Our light CNN model aims to squeeze maximum of accuracy (i.e. functional performance) from a very low number of resources, thus enabling the efficient use of low-energy computing platforms like SoC (systems of chips). For instance, while in [7] several hundreds of convolution layers are needed to get 99.51% accuracy on the handwritten recognition problem, our solution would require only 2 such layers at a very low complexity (around 80k parameters) while achieving just slightly lower 99.22% accuracy.

It is now widely recognized [8] that using quantization, particularly the binary one, will dramatically reduce computational complexity and need for resources, particularly for low-power [9] and FPGA implementations [10]. While various solutions are now proposed for traditional CNN training, adapted for binary quantization [11], we pursue the approach first presented in [12] due to its simplicity and potential for further improvements. It basically defines a nonlinear expander (where convolution layers with binary kernels can be employed instead local receptive fields in [12]) and trains an extreme learning machine (ELM) in the resulted expanded space. In [13] we introduced the B-CONV CNN architecture, and investigated its potential. The implementation was tailored to GPU computation using the Keras/Tensorflow (https://keras.io/) environment. One drawback found then was the relatively large time needed for ELM training in the loop for selecting the best binary kernels. One training epoch lasted from tens to hundreds of seconds. As described in Section II, the new implementation, based on Chainer [14] and Cupy [15] is 2 to 3 orders of magnitude faster on the same platform (Google Colaboratory[1]). Among the many available machine-learning platforms [16], it turns out that for ELM training the Chainer/Cupy platform achieves dramatic speed-ups of 60-1000 times.

Our work expands preliminary results reported in [17] while adding facilities to save the model in a standardized Keras/Tensorflow format and the possibility to further train the optimized model traditionally (using Keras optimizers) to improve accuracy. This addition also allows the use of the entire set for training since, as noted in [17], due to limited GPU-RAM the fast ELM training should be done with a reduced fraction of the training set. The entire framework is provided in [18] and Section III details on the set of tools included in the framework and their use to optimize and save LB-CNN models. Section IV presents examples on optimized

---

[1] https://colab.research.google.com/

LB-CNN models for various datasets. Section V collects conclusions. One of the most important conclusion is that for small to medium sized datasets the LB-CNN can be easily optimized to offer near state-of-the-art accuracies with very low complexity and all the advantages for hardware implementations. Particularly 100% accuracy was achieved for a widely known face recognition dataset.

## II. LB-CNN ARHITECTURE AND ITS IMPLEMENTATION

The LB-CNN architecture is basically the same as BCONV-ELM[2] introduced in [13] and depicted in Fig.1. It is a particular case of the NL-CNN model described in [19] where the nonlinearity index *nl=(1,1)*.

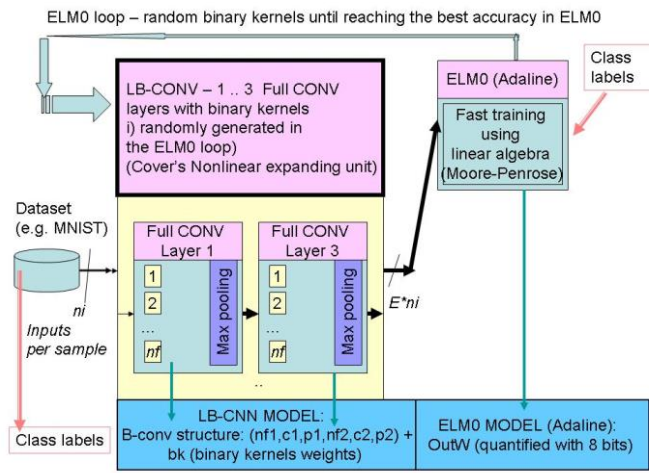

Fig. 1. The structure of the LB-CNN model. Only ELM0 is trainable.

However, here the implementation is considerably different exploiting the deep-learning framework Chainer [14]. Up to three full convolution macro-layers (each including convolution and max-pooling) may be built and applied to the datasets organized in the "channels last" image format. The implementation of a full convolution layer is given next:

```
19  import cupy as cp
20  import chainer
21  from chainer.functions import convolution_2d, depthwise_convolution_2d, relu, flatten, max_pooling_2d, reshape

27  def one_full_layer(net,nf,str,typc,ker,bias):
28  # CONVOLUTIONAL + MAXPOOL layer (Nonlinearity achieved in Max-pool)
29      net=depthwise_convolution_2d(net, ker, b=bias, pad=1)
30      net=max_pooling_2d(net,ksize=4,stride=str, pad=1, cover_all=True)
31      return net
```

The result is a flattened feature vector providing the nonlinearly expanded space of the Cover theorem [20]. Training is done in this expanded space using an Extreme Learning Machine (ELM). The ELM code is a revised version of the one [3] in [21] where the CUPY library [15] was conveniently exploited to run the specific computations on GPU. It is important to note that using CUPY allows a very high productivity, no major changes in the original code being needed, since many basic linear algebra functions in NUMPY have their identical counterpart in CUPY. Although one can use a hidden layer in the ELM, herein we applied the feature vector directly to the Adaline output layer thus reducing drastically the number of parameters. The structure is equivalent to Adaline (linear perceptron) and will be denoted as ELM0 (0 – indicating the number of hidden neurons). The weights are trained using the specific ELM single-epoch approach where some linear algebra operations are performed, as indicated in the implementation bellow:

```
61  # Moore - Penrose computation of output weights (outW) layer
62  # H contains the batch of all Ntr inputs samples (feature-vectors)
63  # Y is the batch of Ntr desired outputs (in categorical format)
64  if n_inputs<Ntr:
65      print('LLL - Less inputs than training samples')
66      outW = cp.linalg.solve(cp.eye(n_inputs)/C+cp.dot(H,H.T), cp.dot(H,Y.T))
67  else:
68      print('MMM - More inputs than training samples')
69      outW = cp.dot(H,cp.linalg.solve(cp.eye(Ntr)/C+cp.dot(H.T,H), Y.T))
70  return outW
```

Note that cp.dot(..) is a simple direct replacement of the np.dot() where cp is the abbreviation for CUPY and np for NUMPY. Although very similar and easy to replace, cp.dot() runs on GPU instead, giving important acceleration benefits. The same stands for all other linear algebra functions.

## III. THE OPEN SOURCE FRAMEWORK AND ITS USE

The framework for training and optimizing LB-CNN models is available in [18]. It is organized as a Jupyter notebook with cells written in Python. The role of the cells is sketched in Fig. 2.

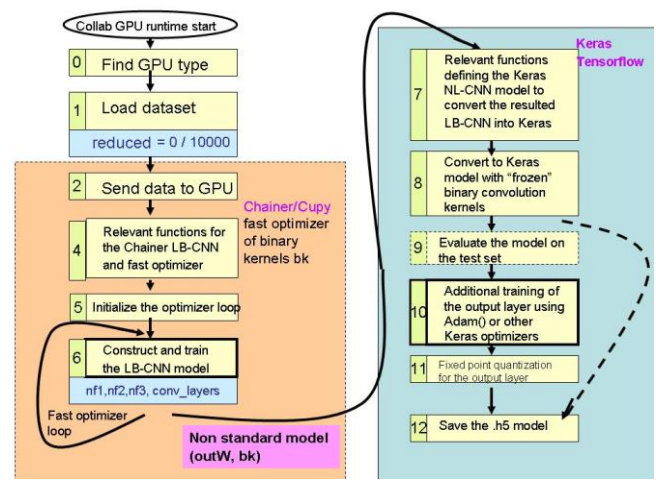

Fig. 2. The flow of using the framework for generating optimized LB-CNN models.

**Cell 0** is used only to report the GPU available on the specific computing platform. **Cell 1** is used to load the dataset associated to the specific problem. Besides **Cell 1a** and **Cell 1b** the user may add similar **Cell 1x** cells to load specific datasets. Note the possibility to use a reduced fraction of the training parameters, as needed by the Chainer/Cupy implementation. **Cells 2,4,5 and 6** implement the fast binary kernel optimizer relying on fast ELM0 training, as follows: **Cell 2** has the role to move all training / test dataset on GPU's memory for further processing. **Cell 4** defines all relevant functions for the LB-CNN model: ELM0 training and prediction (although support exists for one-hidden layer ELMx, as well), the LB-CNN model structured according to the NL-CNN setup [19] and a function to generate randomly the binary kernels **kers**. **Cell 5** is used to initialize the loop process where optimal binary kernels **bk** are sought by running **Cell 6** multiple times and keeping evidence of the best accuracy performance. The structure of the model is defined in **Cell 6** (**conv_layers** – convolution layers, **nf1 nf2 nf3** – number of filters per each convolution layer and nonlinearity index [19], **nl=(1,1)** in most experiments) and each running of it reveals dynamic and structural performance as well (ELM training time, number of bits for parameters, etc.). To explore various random kernels, the variable **use_stored=0** and **Cell 6** runs as many times (trials) as needed to observe an increased performance. Note that GPU-RAM overload problems may occur running **Cell 6**, when very large expansions due to large number of filters are chosen. In such case, one needs to restart the runtime. Running the cell after **Cell 6** gives the best and the lowest performance during the loop training. One needs to re-run **Cell 6** with **use_stored=1** indicating that the best kernels bk are used. The optimized LB-CNN is now available in a non-standard format (the list **bk** of the binary kernels and the output layer weights array) **outW**, in a fixed-point representation with up to 8 bits.

---

[2] Code available as https://github.com/radu-dogaru/LightWeight_Binary_CNN_and_ELM_Keras

[3] https://github.com/radu-dogaru/ELM-super-fast

Further processing expands work in [17] as follows: After running **Cell 7** (defining the LB-CNN model in the Keras/Tensorflow format) **Cell 8** converts the fast optimized non-standard LB-CNN model into a Keras model. Working with a Keras model enables a wide variety of deployment tools that are readily available (for instance, the use of TFLITE library for mobile platforms). One can directly save the model using **Cell 12** for further use with other tools. For instance, using tools in [22] allows the deployment to an FPGA-based SoC platform. Many other tools for deploying Keras models into various platforms are available too.

The most interesting action next is to increase accuracy by additional training (running **Cell 10**) with the traditional Keras approach based on "optimizers". The process is much slower (hundreds of seconds) but is worth applying it, as demonstrated in Section IV, particularly for large datasets. While binary kernels are already optimized from **Cell 6**, the accuracy may be still improved, training the output layer with the full dataset using **Cell 10** (ensuring the batch-based training which is not exhausting the GPU memory). In the case of using the full training set one needs to run **Cell 1** again, now choosing **reduced**=0. Additional fixed-point quantization is achieved by running **Cell 12**.

## IV. EXPERIMENTAL RESULTS

### A. Training speed of the binary kernels optimizer

Herein we consider the widely known medium size dataset (MNIST)[4] and compare two similar implementations in order to evaluate the efficiency of the software platforms and libraries. A 2-layers LB-CNN is built with *nf1*=16 filters and *nf2*=20 filters. Pooling size is 4 for all layers. Strides = 2 on pooling layers, in order to down-sample by a factor of 2 the images in each convolution stage. Only the first 20k (of all 60k) samples were used for training to avoid memory overflow on the available GPU. Both implementations were run on the Google Colab cloud platform with a GPU runtime activated. The 2-layer convolutions expand the input image (28x28 pixels for MNIST to a relatively large feature vector with size 15680 – i.e. expansion factor *E=20*) in order to improve accuracy. Such an input size makes ELM training longer. Using several trials, as seen in Fig.3, best accuracy observed for both implementation was 98.76 %. The results are summarized in Table I.

---
[4] http://yann.lecun.com/exdb/mnist/

Note the impressive increase in performance when using the CUPY library for ELM, **with a reduction of the training time of more than 60 times !** Moreover, the implementation of the convolution layers are much, much faster using the CHAINER framework, with **speed-ups of 1800 times ! Latencies are also 1000 times better.**

TABLE I.  SPEED COMPARISON BETWEEN LB-CNN (CHAINER) AND BCONV-ELM (KERAS) FOR A MEDIUM SIZE TRAINING PROBLEM

| Implementation | Parameters | | |
|---|---|---|---|
| | *convolutional layers computing (s)* | *ELM0 training (s)* | *Latency for prediction (ms)* |
| Chainer/Cupy | 0.0047 | 3.54 | 0.005 |
| [13] Keras/Tensorflow | 9.24 | 243 | 5.3 |

Definitely, for this kind of architecture, using the Chainer/Cupy framework is the best option in order to pursue "in the loop" optimization of the binary kernels. It offers as "side-effect" the fastest ELM training known so far. Herein randomly generated binary kernels are generated in each loop iteration. However, other solutions such as genetic algorithms or swarm intelligence methods may be also considered to generate kernels in the optimization loop. Note the very fast training dominated mostly by the ELM training (where basic linear algebra functions are employed). With lower expansion factors *E* these times are becoming insignificant. This is a major breakthrough, achieved by simply employing a well optimized linear algebra and convolution library targeting the GPU platform.

### B. Choosing the best model by training in the loop

Given the very fast ELM training, one can run as many trials as needed (50 in the example bellow) in a loop, in order to identify the best solution. Fig. 3 shows the accuracies observed in consecutive trials indicating that best solution 98.76% is achieved in less than 20 trials.

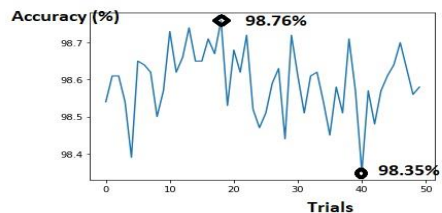

Fig. 3. Accuracy as obtained in 50 consecutive trials (total time 150 seconds) for the MNIST problem.

As noted, the average value is around 98.6%, so accepting a small performance decrease, even one single trial (or better 5) would be sufficient. Consequently, one may tune other hyper-parameters of the model (number of convolution layers, filters, etc.) for best accuracy, in reasonable amounts of time.

In the next, several widely known datasets will be considered and indicate the best solution, giving a comparison between several solutions in terms of accuracy (on the test set), and number of bits for the parameters (considering that the output Adaline layer has weights represented as 8 bits integers). The neural model implemented is the NL-CNN [19] with $nl=(1,1)$ nonlinearity index but with binary weights in convolutional layers and depth-wise convolutions. Unlike in the case of the fully-trainable NL-CNN our experiments show that $nl=(1,1)$ i.e. the L-CNN model [13] and depth-wise instead normal convolution would give the best performance in the case of restricted binary weights. The number of filters per layer are the only changeable parameters. In a list of 3 for instance (2,3,-) means 2 filters on the first layer, 3 filters on the second layer and the absence of the third layer. The only drawback for the depth-wise convolution is that it multiplies the number of image channels at each stage, thus explaining the low number of filters. Since all data has to be stored on GPU RAM this gives an important limitation in choosing larger numbers of filters. On smaller or medium sized training samples, the above effect is limited. Summarizing, the following tables report some of the best solutions, with the best solution emphasized on yellow rows and being stored as .h5 model in [18]. The columns 2 and 3 in the tables report the best performance for the ELM0 output model (fast optimizer). For large datasets (with more than 10000 samples) reduced sets with 10000 samples were used to avoid the GPU RAM overflow. The 4 th and the 5-th column report improved results when the Keras-based Adaline (output layer) training + quantization is used, using the entire training dataset (at the expense of much larger training times)

C. *Improved solution for MNIST dataset*

Several trials were done to identify the best LB-CNN solution for MNIST. The MNIST problem is a widely known one, and for a two-layers CNN with trainable convolutional kernels the accuracy on the test set reaches 99.11% accuracy in a setup with a similar number of filters. The next table summarizes our results using LB-CNN. Best solution is on the yellow row.

TABLE II. MODELS TO IDENTIFIY THE MOST EFFICIENT MNIST LB-CNN MODEL

| Model nf1,nf2,nf3 | Accuracy ELM0 (%) | Cupy Train time | Accuracy after Keras train | Keras Train time | Params (k-bits) Conv+ELM |
|---|---|---|---|---|---|
| 40,2,2 | 98.11 | 0.03 s | 98.59 | 211 s | 2.52+204.8 |
| 40,4,- | 98.49 | 3.7 s | 99.22 | 356 s | 1.8+627 |

As discussed in [13], one convolution layer does not offer a very good performance, three layers in principle can do, but expanding too much the input raises memory overflow so for this dataset given the GPU memory constraints a two-layer convolution is the best compromise, achieving 98.49% accuracy with a small number of parameters. As indicated, using additional output layer training in the Keras framework raises the accuracy up to 99.22% not very far from state of the art fully-trainable non-binary solutions (with about 99.8%). Similar problems occur for the more difficult CIFAR10 dataset. The best model achieved 67.35% accuracy (the .h5 model saved in [18]) using a model with $nf1=10$ and $nf2=6$ filters with a size of 1.89+921 kbits).

D. *Improved solution for the German traffic sign set*

The "German traffic sign set" [23] (GTSRB) contains a number of 43 classes with color pictures (32x32 sized, here) representing European traffic signs. There are 34799 samples in the training set. This set is relevant for various automotive applications. State of the art performance gives around 99% accuracies but most low complexity models report lower values, as expected (under 91% in [24] table 7, for the low complexity EFF neural net model). The most significant results using LB-CNN are depicted in Table III.

TABLE III. MODELS TO IDENTIFY THE MOST EFFICIENT GERM-TRAFFIC SET LB-CNN MODEL

| Model nf1,nf2,nf3 | Accuracy ELM0 (%) | Cupy Train time | Accuracy after Keras train | Keras Train time | Params (k-bits) Conv+ELM |
|---|---|---|---|---|---|
| 13,-,- | 89.2 | 6.62 s | 90.8 | 345 s | 0.351+3434 |
| 20,2,- | 90.37 | 3.2 s | 91.28 | 221 s | 1.62+2641 |

Note that in this case, best accuracies are achieved in a two convolution layers model. As in the previous cases, increasing the number of filters can still improve accuracy but our reported solution is limited by the available GPU memory resources (around 12Gbytes). Given the very low complexity, 91.28% accuracy is a good result, in accord to similar low complexity solutions reported in the literature.

E. *Improved solution for the ORL dataset*

This dataset [5] is among the first face recognition sets introduced and is widely known. For many years, various machine learning solutions were tried on it, reporting accuracies under 100%. A trainable CNN (the L-CNN solution in [13]) was able to reach a near optimal 99.33% accuracy on the test set. On the other hand, in [25] where many solutions were compared for the ORL dataset the authors proposed a quite complicated solution (ResNet50 +SVM) in order to achieve maximal 100% accuracy. As seen, the LB-CNN achieves 100% accuracy for a very light architecture, as indicated in the next table. Face images of 64x64 pixels (gray, reduced size) were used in our experiment.

TABLE IV. MODELS TO IDENTIFY THE MOST EFFICIENT LB-CNN MODEL FOR THE ORL FACE RECOGNITION SET

| Model nf1,nf2,nf3 | Accuracy ELM0 (%) | Cupy Train time (s) | Accuracy after Keras train | Keras Train time | Params (k-bits) Conv+ELM |
|---|---|---|---|---|---|
| 4,-,- | 93.33 | 0.0017 | 93.33 | 6s | 0.036+1310 |
| 4,4,- | 99.33 | 0.0014 | 99.33 | 6s | 0.18+1310 |
| 5,4,- | 100 | 0.0014 | 100 | 6s | 0.225+1638 |

Quite interesting, in this case, 2 convolution layers suffice to improve performance and the resulted solution is rather light. As for any other problems with small training datasets the best result is obtained without the need of additional Keras training in the output layer.

F. *Improved solution for the reduced VGG dataset*

The VGG dataset [6] is here reduced to a selection of 6 particular individuals (among thousands of identities) in order to reduce the dimension of the set. There are now 1304 training samples belonging to these classes. Each sample (image) has a reduced size of 64x64 pixels. Training an L-CNN [13] (simple light-weight, trainable CNN) hardly gives

---

[5] http://cam-orl.co.uk/facedatabase.html

[6] https://www.robots.ox.ac.uk/~vgg/data/vgg_face/

90% accuracy, while results for up to 95% were recently reported [26] with a more carefully designed light-weight architecture, but for 5 classes. As shown in the next table best performance is achieved in a 3-layes structure. Possibly deeper LB-CNN would do it better.

TABLE V. MODELS TO IDENTIFIY THE MOST EFFICIENT LB-CNN MODEL FOR THE VGG6 FACE RECOGNITION SET

| Model nf1,nf2,nf3 | Accuracy ELM0 (%) | Cupy Train time (s) | Accuracy after Keras train | Keras train time (s) | Params (k-bits) Conv+ELM |
|---|---|---|---|---|---|
| 20,4,4 | 87.38 | 0.0024 | 87.38 | 34 | 11.3+2949 |
| 10,4,2 | 87.69 | 0.0024 | 88 | 34 | 3.5+737 |

Given the difficulty of this problem, an accuracy of 88.00% is still a good result traded off for the very low complexity of the model.

V. CONCLUSIONS

A framework for optimizing and training very light CNN models with binary convolution kernels is described. It was tested on the freely available Google Collaboratory platform and is available at [18]. Binary kernels and fixed-point quantization of the simple linear output classifier ensures light-weight implementations, particularly where low-power hardware implementations are sought (FPGA, SoC or dedicated hardware). The proposed framework combines the excellent speed for ELM training of the output layer offered by the Chainer/Cupy framework with the model versatility and wide availability of deployment tools given by the Keras/Tensorflow models. The resulting model is available in the standard .h5 format and can be further deployed on a wide variety of Edge AI platforms to serve in various industrial applications based on image recognition. Only 3 types of layers are used in our model: namely, **depthwise-convolution**, **max-pooling**, and **dense** thus making very easy the process of defining similar layers on a variety of platforms for deployment.

For small training set the fast optimizer based on ELM training of the output layer and random choices for binary convolutional kernels works fine, allowing a rapid optimization of the model hyperparameters (number of convolutional layers and filters). Among many machine learning libraries we found that Chainer/Cupy is the best choice to implement the fast optimizer due to the very efficient ELM training obtained in this case (orders of magnitude faster than using other frameworks on the same platform). For larger datasets, the fast optimizer relying on Chainer framework may have problems due to the large GPU RAM needed and consequently the training is done in this case on a reduced set of training samples, still requiring no more than 6 seconds per trial. But due to converting the resulting model into the Keras/Tensorflow format one can further improve the accuracy of the mode while maintaining its complexity doing additional output layer training while keeping the optimized binary kernels frozen.

Although performance is maximal for some small datasets (for instance 100% accuracy on the ORL face recognition problem) it is generally lower than state of the art for difficult problems such as CIFAR10 but comparable with other similar TinyML [27] solutions where low complexity is an important issue. Improvements in accuracies are the subject of future work, some possible solutions being the use of committees (based on several LB-CNNs) as suggested in [28]. Also, including some advanced optimization techniques (swarm optimization, genetic algorithms) to replace random generation of binary kernel in the ELM-training loop may increase accuracy at the expense of speed. In terms of parameters, in all studied cases, the most parameters are allocated in the output layer, and a very low number of bits are allocated to the otherwise computationally intensive convolution layers. Consequently, the use of binary weights substantially reduces the number of devices (logical gates) to be allocated in the stage of computing convolutions with important effects on the power consumption of the embedding platform.